\documentclass[lettersize,journal]{IEEEtran}
\usepackage{amsmath,amsfonts}
\usepackage{algorithmic}
\usepackage{algorithm}
\usepackage{array}
\usepackage[caption=false,font=normalsize,labelfont=sf,textfont=sf]{subfig}
\usepackage{textcomp}
\usepackage{stfloats}
\usepackage{url}
\usepackage{verbatim}
\usepackage{graphicx}
\usepackage{cite}
\usepackage{amsmath,amssymb,amsfonts,bm,amsfonts}
\usepackage{algorithmic}
\usepackage{balance}
\usepackage{textcomp}
\usepackage{newfloat}
\usepackage{listings}
\usepackage[table]{xcolor}

\usepackage{algorithm}
\usepackage{multirow}
\usepackage{hyperref}
\usepackage{url}
\usepackage{booktabs}
\usepackage{colortbl,bm,amsfonts}
\usepackage{amsmath}
\usepackage{array}
\usepackage{enumitem}
\usepackage{caption}
\usepackage{tabularx}
\usepackage{pifont}
\usepackage[T1]{fontenc}

\usepackage{amsmath,amsfonts,bm}
\usepackage{algorithm,amsmath,amssymb,bm,amsthm}
\usepackage{algorithmic}
\usepackage{enumitem}
\newtheorem{theorem}{Theorem}[section]

\newtheorem*{lemma*}{Lemma}

\def\eqref#1{equation~\ref{#1}}

\def\1{\bm{1}}

\def\vb{{\bm{b}}}

\def\vf{{\bm{f}}}

\def\vw{{\bm{w}}}
\def\vx{{\bm{x}}}
\def\vy{{\bm{y}}}
\def\vz{{\bm{z}}}

\DeclareMathAlphabet{\mathsfit}{\encodingdefault}{\sfdefault}{m}{sl}
\SetMathAlphabet{\mathsfit}{bold}{\encodingdefault}{\sfdefault}{bx}{n}

\def\gB{{\mathcal{B}}}
\def\gC{{\mathcal{C}}}
\def\gD{{\mathcal{D}}}
\def\gE{{\mathcal{E}}}

\def\gG{{\mathcal{G}}}
\def\gH{{\mathcal{H}}}

\def\gL{{\mathcal{L}}}

\def\gN{{\mathcal{N}}}

\def\gP{{\mathcal{P}}}

\def\gT{{\mathcal{T}}}

\def\gV{{\mathcal{V}}}
\def\gW{{\mathcal{W}}}

\def\gY{{\mathcal{Y}}}

\newcommand{\E}{\mathbb{E}}

\def\eg{\emph{e.g}.} 
\def\ie{\emph{i.e}.}

\definecolor{LightCyan}{rgb}{0.88, 0.96, 0.92}
\definecolor{cvprblue}{rgb}{0.21, 0.74, 0.49}
\newcommand{\fix}[1]{\textcolor{black}{{#1}}}
\hypersetup{
    colorlinks=true,
    citecolor=blue,
    linkcolor=blue,
    urlcolor=black
}

\newcommand{\cmark}{\ding{51}}

\usepackage{color}
\def\BibTeX{{\rm B\kern-.05em{\sc i\kern-.025em b}\kern-.08em
    T\kern-.1667em\lower.7ex\hbox{E}\kern-.125emX}}

\begin{document}

\def\method{COUPLE}
\newcommand{\paratitle}[1]{\vspace{1ex}\noindent\emph{\textbf{#1}}}
\title{Cross-Domain Diffusion with Progressive Alignment for Efficient Adaptive Retrieval}

\author{
Junyu Luo, Yusheng Zhao, Xiao Luo, Zhiping Xiao, Wei Ju, \\
Li Shen, Dacheng Tao,~\IEEEmembership{Fellow,~IEEE}, Ming Zhang
\IEEEcompsocitemizethanks{
\IEEEcompsocthanksitem Corresponding to Xiao Luo, Zhiping Xiao and Ming Zhang.
\IEEEcompsocthanksitem Junyu Luo, Yusheng Zhao, Wei Ju, Ming Zhang are with State Key Laboratory for Multimedia Information Processing, School of
Computer Science, Peking University, Beijing, China. (e-mail: luojunyu@stu.pku.edu.cn,  yusheng.zhao@stu.pku.edu.cn, juwei@pku.edu.cn, mzhang\_cs@pku.edu.cn)
\IEEEcompsocthanksitem Xiao Luo is with Department of Computer Science, University of California, Los Angeles, USA. (e-mail: xiaoluo@cs.ucla.edu)
\IEEEcompsocthanksitem Zhiping Xiao is with Paul G. Allen School of Computer Science \& Engineering, University of Washington, Seattle, USA. (e-mail: patricia.xiao@gmail.com)
\IEEEcompsocthanksitem Li Shen is with JD Explore Academy. (e-mail: mathshenli@gmail.com)
\IEEEcompsocthanksitem Dacheng Tao is with College of Computing \& Data Science, Nanyang Technological University, Singapore. (e-mail: dacheng.tao@ntu.edu.sg)
}

}



\maketitle

\begin{abstract}

Unsupervised efficient domain adaptive retrieval aims to transfer knowledge from a labeled source domain to an unlabeled target domain, while maintaining low storage cost and high retrieval efficiency. However, existing methods typically fail to address potential noise in the target domain, and directly align high-level features across domains, thus resulting in suboptimal retrieval performance. 
To address these challenges, we propose a novel 
Cross-Domain Diffusion with Progressive Alignment method~(\method{}).
This approach revisits unsupervised efficient domain adaptive retrieval from a graph diffusion perspective, simulating cross-domain adaptation dynamics to achieve a stable target domain adaptation process.
First, we construct a cross-domain relationship graph and leverage noise-robust graph flow diffusion to simulate the transfer dynamics from the source domain to the target domain, identifying lower noise clusters. We then leverage the graph diffusion results for discriminative hash code learning, effectively learning from the target domain while reducing the negative impact of noise. Furthermore, we employ a hierarchical Mixup operation for progressive domain alignment, which is performed along the cross-domain random walk paths.
Utilizing target domain discriminative hash learning and progressive domain alignment, \method{} enables effective domain adaptive hash learning. 
Extensive experiments demonstrate \method{}'s effectiveness on competitive benchmarks.

\end{abstract}


\section{Introduction}

\IEEEPARstart{A}{pproximate} nearest neighbor~(ANN)~\cite{li2019approximate,zhen2019deep} search, which aims to efficiently find data samples in a dataset that are close to a given query sample within an acceptable margin of error, has garnered significant attention due to its wide range of applications, \eg, image retrieval~\cite{10440048,10382463}, search engines~\cite{ji2019efficient}, recommender system~\cite{tan2020learning} and retrieval-augmented generation~(RAG)~\cite{agentsurvey2025,zhao2024retrieval}. Hash-based ANN retrieval~\cite{yan2020deep,mm23-graphhash-shen,cui2024effective,10460427,10418867} offers higher efficiency and lower storage costs by replacing computationally expensive pairwise distance calculations with bit-wise XOR and bit-counting operations~\cite{wang2020hash}. As data and knowledge scale up, the field has gained growing interest from the research community. 
With the development of deep learning techniques, deep hashing is proposed to map high-level features to compact hash codes while preserving semantic similarity~\cite{zhu2023multi}.
Existing deep hash learning techniques include supervised methods~\cite{mm19-hash-yan,mm20-hash-zhan,9953581,mm23-hash-lu} and unsupervised methods~\cite{mm20-unsup-hash-jin,wang2022learning,mm23-unsup-hash-song,mm22-unsup-hash-li,9543564,xiao2023dlbd}. Supervised hashing methods leverage label information to integrate deep hash learning into representation learning, usually achieving superior retrieval performance compared to unsupervised hashing methods. These supervised methods can generate highly discriminative hash codes, leading to encouraging results in efficient information retrieval.

However, supervised methods typically assume that the training and query data follow the same distribution, which may not hold in real-world scenarios. In practice, models often encounter out-of-distribution~(OOD) problems. For instance, we may wish to retrieve studio-captured product images using mobile-captured images, or enable cross-retrieval between different generative models~(\eg, DALL-E~\cite{dalle}, Stable Diffusion~\cite{rombach2022high}). 
To address this OOD issue, in this paper, we focus on unsupervised domain adaptive hashing, aiming to enhance the performance of a hash retrieval model pre-trained on a labeled source domain using unlabeled target domain data. In recent years, domain adaptive hashing methods~\cite{mm21-adahash-xia,huang2021domain,mma22-adahash-shi,wang2024idea} have gradually gained attention from researchers, focusing on addressing the issues of \textit{target label scarcity} and \textit{large domain discrepancies}. \textit{To solve data scarcity}, existing methods~\cite{venkateswara2017deep,mm21-adahash-xia,wang2023toward} typically leverage pseudo-label learning to extract knowledge from unlabeled target domain data. \textit{To solve large domain discrepancies}, existing methods~\cite{long2018deep,he2019one,ma2024discrepancy} often employ adversarial learning or domain discrepancy minimization for domain alignment. Both directions have made notable progress on domain adaptive hashing retrieval.

Unfortunately, domain adaptive hashing retrieval still faces two unresolved challenges that hinder further performance improvements. \textit{\textbf{(1) Noise in the target domain remains inadequately addressed.}} Domain discrepancies can introduce noise into the target domain, causing the model to become over-confident about the target samples. This over-confidence can lead to biased or incorrect outputs, resulting in the ineffectiveness of self-learning techniques like pseudo-label learning~\cite{lee2013pseudo}. The resulting performance decline introduces more biased outputs, creating a positive feedback loop between noisy outputs and model overconfidence. This cycle significantly hinders effective knowledge mining in the target domain. \textit{\textbf{(2) Domain discrepancies are not yet effectively eliminated.}} Existing methods~\cite{huang2021domain,mma22-adahash-shi,wang2024idea} typically perform domain discrepancy minimization or adversarial learning on all source and target samples in the semantic-rich high-level feature space. 
However, real-world domain differences are hierarchical, encompassing both high-level disparities~(\eg, semantic aspects) and low-level variations~(\eg, image style differences). Consequently, attempting to eliminate domain discrepancies solely in high-level spaces hinders effective domain alignment during domain adaptive retrieval. 

To address the above challenges, we introduce a \textbf{C}ross-D\textbf{o}main Diff\textbf{u}sion with \textbf{P}rogressive A\textbf{l}ignm\textbf{e}nt method~(\textbf{\method{}}) from a graph diffusion perspective. 
This approach simulates domain transfer dynamics, employing a divide-and-conquer strategy. 
\method{} prioritizes domain adaptation for \textit{early adopters} with lower noise levels. For data with higher noise, \method{} utilize hierarchical Mixup for progressive domain adaptation. This method effectively balances the adaptation process across varying noise levels. 
Specifically, we first construct a cross-domain relationship graph based on mutual nearest neighbors between source and target domain samples. 
Then, we perform cross-domain graph flow diffusion to detect \textit{early adopters}, which are less affected by the model's overconfident and biased predictions, exhibiting lower noise levels. \method{} could effectively identifies sub-cliques with better connectivity and higher confidence within the target domain. We theoretically prove the robustness of our method against target domain noise~(see Section~\ref{sec:diffusion}).
We prioritize these \textit{early adopters} due to their lower noise levels, utilizing cross-domain discriminative learning to extract knowledge from the target domain. 
To maximize the utility of target domain data, we implement a progressive Mixup mechanism and consistency learning for higher-noise data. Specifically, to address the limitations of direct high-level feature alignment, we introduce a hierarchical Mixup mechanism based on the cross-domain relationship graph, facilitating progressive domain alignment.
This progressive Mixup effectively reduces the learning difficulty of domain alignment. 
By integrating these processes, our method enables effective progressive domain adaptation hashing. We conduct extensive comparisons with various state-of-the-art methods on a series of benchmarks, demonstrating the effectiveness of our approach.
The contributions of our work can be highlighted as follows:
\begin{itemize}[leftmargin=*]
    \item \paratitle{New Perspective.} We study unsupervised domain adaptation hashing from a graph diffusion perspective for efficient cross-domain retrieval. To the best of our knowledge, this is the first attempt to employ graph diffusion in domain adaptation tasks.
    \item \paratitle{New Method.} We propose a new method \method{} for domain adaptive retrieval. \method{} achieves noise-robust target domain discriminative learning and progressive cross-domain alignment through neighborhood relationship graphs and graph flow diffusion.
    \item \paratitle{Sate-of-the-art Performance.} Comprehensive experiments on benchmark datasets demonstrate that our \method{} achieves more effective hashing learning and retrieval capabilities compared to the state-of-the-art baselines.
\end{itemize}

The remainder of this paper is structured as follows. Section~\ref{sec::related} reviews relevant research advancements. Section~\ref{sec::method} introduces the research problem and proposed \method{}. Section~\ref{sec::experiment} presents experimental analysis and discussion of \method{}. Finally, Section~\ref{sec::conclusion} concludes the paper.

\begin{figure*}[t]
    \centering
    \includegraphics[width=0.98\textwidth]{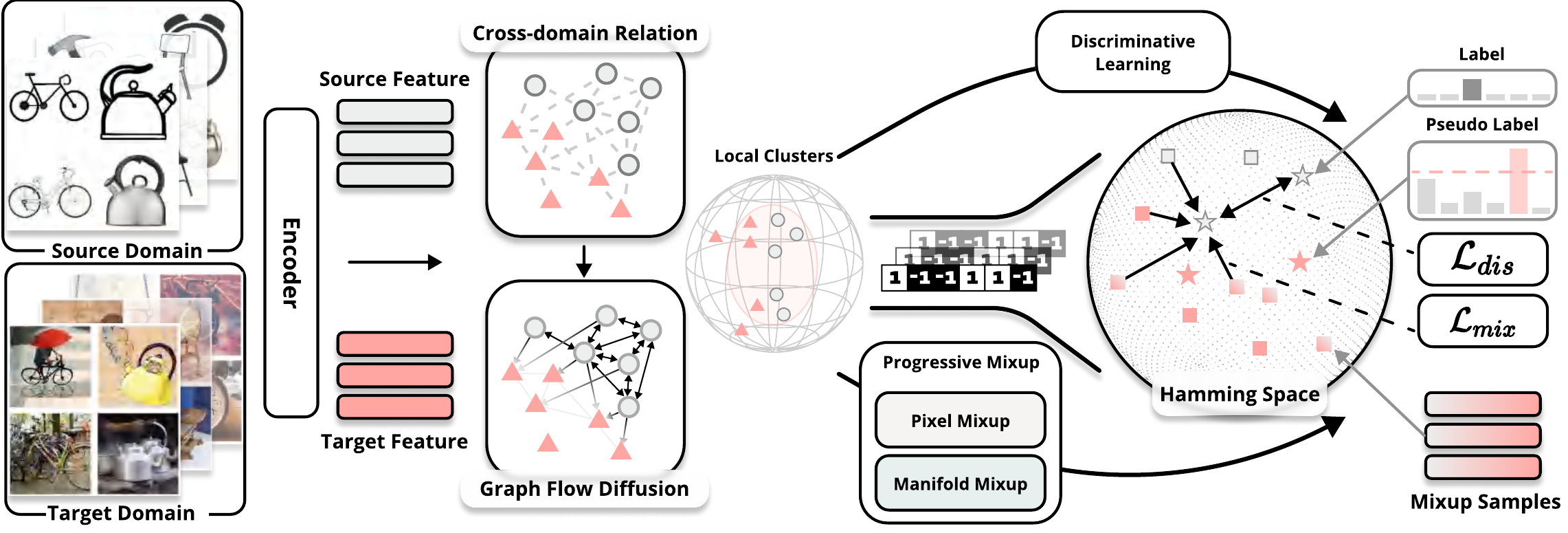}
    \caption{Overview of \method{}. Our objective is to learn domain-adaptive hash codes for the target domain. Specifically, we leverage cross-domain diffusion to simulate the transfer dynamics and explore the target domain robustly. Furthermore, we perform hierarchical Mixup learning along cross-domain random walk paths to achieve progressive domain alignment.}
    \label{fig:fig2}
\end{figure*}

\section{Related Work}
\label{sec::related}

\subsection{Deep Hashing}
Deep hashing has enabled efficient information retrieval~\cite{hash-survey,Wang_2023_CVPR,Chen_2024_CVPR}, with applications in cross-modal retrieval~\cite{Doan_2022_CVPR,mm20-hash-zhan,9670722,9782584,8543225,10502297,sun2024dual}. \fix{Recent advances have demonstrated its effectiveness in hierarchical consensus learning~\cite{sun2023hierarchical} and zero-shot sketch-based retrieval scenarios~\cite{wang2023cross}.} These methods significantly reduce storage and computational requirements.
Existing approaches include supervised methods~\cite{mm20-hash-zhan,9953581,mm23-hash-lu} and unsupervised methods~\cite{mm20-unsup-hash-jin,wang2022learning,mm23-unsup-hash-song,mm22-unsup-hash-li,9543564,xiao2023dlbd}, with the former utilizing label information, and the latter employing self-supervised learning and semantic structure similarity reconstruction. Supervised methods, thanks to the rich label information, exhibit superior performance to other baselines, generally being categorized into similarity-based or label-based methods. Similarity-based supervised methods~\cite{mm19-hash-chen,wang2023deep} utilize pairwise loss and ranking loss, while label-based supervised~\cite{wang2023toward} methods enhance hash codes using classification loss and map them to Hamming space using Hadamard matrices or Bernoulli distributions. Despite remarkable progress on standard benchmarks, potential distribution shifts from source to target domain in real-world applications can significantly impact the retrieval accuracy. Therefore, this paper investigates domain-adaptive deep hashing to address such practical challenges faced in real-world scenarios.

\subsection{Unsupervised Domain Adaptation}
Unsupervised domain adaptation~(UDA) is a crucial research topic for practical applications~\cite{ju2024surveydataefficientgraphlearning}, aiming to transfer knowledge from a labeled source domain to an unlabeled target domain~\cite{long2018conditional,he2022secret,rna2024} under the OOD problem. Existing methods include domain alignment, discriminative learning, and adversarial learning approaches. Domain alignment methods~\cite{Lee_Batra_Baig_Ulbricht_2019,gala2024} typically measure and minimize the discrepancy between domains using metrics such as Wasserstein distance. Discriminative learning methods~\cite{wei2021metaalign,xiao2021dynamic} often employ pseudo-labeling~\cite{lee2013pseudo} and consistency learning to leverage unlabeled target data for model adaptation. Adversarial learning-based methods~\cite{ganin2016domain}, utilize domain discriminators and gradient reversal layers for knowledge transfer. Despite the success of UDA methods, most of these approaches focus on tasks like image classification and segmentation. In this paper, we focus on applying UDA to efficient data retrieval and explore target domain data from a graph diffusion perspective. To the best of our knowledge, we are the first to utlize graph diffusion methods to simulate domain adaptation dynamics and achieve noise-robust target domain mining.

\subsection{Domain Adaptive Hashing}
Domain adaptive hashing~\cite{yan2020deep,mm23-graphhash-shen,10418867} has recently gained attention for its ability to leverage source domain data and enable cross-domain information retrieval~\cite{zhang2019optimal}. Early works focused on single-domain retrieval tasks. DAH~\cite{venkateswara2017deep} employs pairwise loss and maximum mean discrepancy to reduce domain disparity. DeDAHA~\cite{long2018deep} introduces adversarial learning for domain alignment. DAPH~\cite{huang2021domain} achieves cross-domain retrieval by minimizing domain differences. DHLing~\cite{mm21-adahash-xia} utilizes memory bank operation for better representation generation. \fix{Zhang et al.~\cite{zhang2023dynamic} propose dynamic confidence sampling and label semantic guidance to enhance domain adaptation.} PEACE~\cite{wang2023toward} and IDEA~\cite{wang2024idea} are recently proposed methods, which obtain domain-invariant hash codes for effective cross-domain retrieval.
In this paper, we introduce \method{}, achieving state-of-the-art performance in domain adaptive retrieval. 
Different with existing methods, \method{} is the pioneer in simulating domain adaptation dynamics and using a divide-and-conquer strategy to combat potential noise in the target domain. For samples with higher noise levels that are challenging to mining, we utilize a hierarchical Mixup mechanism for progressive domain alignment.

\section{Methodology}\label{sec::method}

\subsection{Preliminaries and Overview}\label{sec:pre}

\paratitle{Problem Definition.}
The proposed deep hash learning framework addresses the domain shift problem between a labeled source domain $\gD^{s}=\{( \vx^{s}_i, y^{s}_i )\}^{\gN_s}_{i=1}$ with $\gN_s$ samples and an unlabeled target domain $\gD^{t}=\{ ( \vx^{t}_j ) \}^{\gN_t}_{j=1}$ with $\gN_t$ samples, where both domains share the same label space $\gY = \{1,2,\cdots,C\}$. The objective of domain adaptive hasing retrieval is to learn a deep hash model $\gH$ that extracts binary codes $\vb\in \{-1,1\}^L$ of length $L$ from input images $\vx$. \fix{The similarity between two images is measured by the Hamming distance between their hash codes, defined as $d_H(\vb_i,\vb_j)=\frac{1}{2}(L-\vb_i^T\vb_j)$.} The retrieval system ensures that similar images have appropriate binary codes in the Hamming space and validates its effectiveness in both single-domain and cross-domain scenarios.

\paratitle{Framework Overview.}
This work proposes a new efficient adaptive retrieval approach named \method{}. The framework is shown in Figure~\ref{fig:fig2}. Our \method{} encodes the visual features and then maps latent features to a hash code $\vb$, with a length of $L$, using an MLP $\phi(\cdot)$. Following the previous works, the feature extractor is derived from a visual backbone network by replacing the final projection layer. In formula, we have:
\begin{equation}
\vf = F(\vx)\,, \quad
\vb = sign\left( \phi\left (\vf \right ) \right) \,,
\end{equation}
where $\vf$ denotes the latent feature, $F(\cdot)$ represents the visual encoder to encode the input image $\vx$, $sign(\cdot)$ is the sign function and $\vb$ is the target hash codes.  
Our \method{} incorporates two key components: 
\begin{itemize}[topsep=0pt,leftmargin=*]
\item \textbf{\textit{Domain Diffusion-based Discriminative Learning}}, which simulates the cross-domain transfer dynamics~(Section~\ref{sec:diffusion}) and identifies reliable samples for noise-robust discriminative hash learning~(Section~\ref{sec::hash-learning}). We also prove the robustness of our method under target noise theoretically.

\item \textbf{\textit{Progressive Mixup for Domain Alignment}}, which includes intra-cluster neighbors Mixup and inter-cluster path-based Mixup~(Section~\ref{sec::mixup}). This facilitates effective hierarchical cross-domain alignment, and the learning of comprehensive hash codes.

\end{itemize}

\subsection{Cross-Domain Graph Diffusion}\label{sec:diffusion}

The domain discrepancy leads models trend to over-confident and baised prediction, which introduces noise in the target domain. \fix{Specifically, this noise manifests as incorrect pseudo-labels for target domain samples, a common challenge in unsupervised domain adaptation that significantly impacts model performance.} Previous methods did not adequately address this problem. Our \method{} explores cross-domain data relationships from a graph perspective, then simulates cross-domain transfer dynamics and uncovers the intrinsic patterns within cross-domain data samples. 

\paratitle{Cross-domain Relationship Graph Construction.}
To begin with, we model the relationship graph between source and target domain data using the Mutual Nearest Neighbor~(MNN) graph~\cite{chen2022mutual}. 
Here we connect similarity-based neighbors to bridge the source and target domains coarsely. Following common practice, we select samples that are mutual $k$-nearest neighbors, where $k=3$ as previous methods~\cite{chen2022mutual}. However, it is worth noting that similarity-based nearest neighbors only provide noisy cross-domain relationships. 

The construction of cross-domain relationship graph need the  consideration of the following situations. 
\textit{For cross-domain data,} we retrieve the nearest neighbors of each target domain sample $\vx^s_i$ in the source domain as $\gN^s$. Simultaneously, we perform the same operation for each source domain sample $\vx^t_j$ and obtain $\gN^t$ in the target domain. If $\vx_i^s$ and $\vx_j^t$ are mutual nearest neighbors, \ie, $\vx_i^s \in \gN_t(\vx_j^t)$ and $\vx_j^t \in \gN_s(\vx_i^s)$, an edge is established between them, indicating an MNN pair. 
\textit{For intra-source domain data}, since we have access to their ground-truth labels, instances with the same label are considered MNN pairs.
\textit{For intra-target domain data}, due to the lack of ground-truth labels, we follow the same approach as for cross-domain data. 
This process yields an undirected graph $\gG = (\gV, \gE, \gW)$, where the node set $\gV = \gD_s \cup \gD_t$, the edge set $\gE = \left\{ \{i, j\} \mid \vx_i^s \in \gN_t(\vx_j^t),~ \vx_j^t \in \gN_s(\vx_i^s)\right\}$ consists of all MNN pairs, and the graph is weighted by $\gW$ by the similarity of samples.
The MNN graph effectively models the cross-domain relationships, which guide the subsequent domain adaptation process.

\paratitle{Graph Flow Diffusion.}
Building upon the established relationship graph $\gG$, for the first time, we introduce graph flow diffusion to uncover the intrinsic structure within the data. The motivation is to simulate the dynamics of the domain adaptation process and identify more well-connected communities in the cross-domain relationship graph. Graph flow diffusion is investigated in local clustering~\cite{chen20222,yang2023weighted}. However, this technique remains unexplored in the context of domain adaptation dynamics.
We use cross-domain diffusion process to divide samples with different noise level in target domain, and then conque them with different strategies.

The dynamics of domain adaptation process can be modelled by mass transferring. 
To begin, we initialize the mass for the cross-domain data as,
\begin{equation}
\Delta = 
\begin{cases}
deg(\vx_i), & \text{if } \vx_i \in \gD^s \,, \\ 
0, & \text{if } \vx_i \in \gD^t \,,
\end{cases}
\end{equation}
where $deg(\cdot)$ is the degree of each instance in the cross-domain relationship graph.

Then, we employ the graph flow diffusion process  into cross-domain adaptation. Each node is assigned a flow capacity $T_i$, set to half the average degree of the relationship graph. If a node's flow exceeds its capacity~(\ie, $\Delta_i \geq T_i$), it must distribute the excess flow to its neighbors to satisfy the constraint. Each node allocates its current flow to its neighbors proportionally based on the edge weights. The flow diffusion problem can be summarized in the dual problem as $\ell_2$-norm flow diffusion~\cite{chen20222},
\begin{equation}
\min \frac{1}{2} \sum_{e\in E} \frac{f^2}{\gW}  \quad \text{s.t. } B^Tf+\Delta \leq T \,,
\end{equation}
\begin{equation}
\min x^TLx+x^T(T-\Delta)\quad \text{s.t. } x\geq 0\,,
\end{equation}
where $B$ is the edge incidence matrix, $L=B^T\gW B$ is the graph Laplacian matrix, and $f$ is the solution of the flow diffusion problem. After performing the graph flow diffusion process, we can obtain the well-connected communities $\gC$ from the top $\gamma$ value of mass value $x$ as,
\begin{equation}\label{eq:gamma}
\gC = \arg\max_{\vx_i \in \gD^t}(x_i, \gamma)\,,
\end{equation}
where $\gamma$ is investigated in Sec~\ref{sec:sensitivity} and set default to $50\%$.
$\gC$ is the set that we select to have the lower noise level in the target domain.
In the following part, we will give priority to $\gC$ for effective adaptive hash learning. Before that, we first discuss the noise-robustness of our \method{}. 

\paratitle{Discussion on the Noise-robustness.}
In this part, we discuss the noise-robustness of \method{} in the cross-domain modeling scenario. Due to the discrepancy across domains, the model's prediction for the target domain may turn to over-confidence and bias, resulting in potential noise in the target domain. 
Cross-domain graph flow diffusion aims to identify more reliable confident clusters, considering noise in the target domain, \ie, making samples within confident clusters exhibit higher prediction accuracy.

Let $\gT$ denote the ideal, noise-free target domain set, comprising the model's correct predictions in the target domain. Conversely, $\gN$ represents the set of the model's incorrect predictions, \ie, 
\begin{equation}\label{eq::target-noise}
\gN = \gD^t \setminus \gT \,,
\end{equation}
which is the noise in the target domain. Moreover, $\gC$ in Eq.~\ref{eq:gamma} is the low-noise target domain set we recover.
Then, we can calculate the noise assessment accuracy by:
\begin{equation}
a_1 := \frac{\lvert \gT \cap \gC \rvert}{\lvert \gT \rvert} \,, \quad\quad
a_0 := \frac{\lvert \gN \setminus \gC \rvert}{\lvert \gN \rvert} \,.
\end{equation}
Subsequently, the performance of cross-domain diffusion process can be assessed by the F1 score as,
\begin{equation}
ACC \left( \gC \right) = \frac{
2\lvert \gT \rvert
}{
2\lvert \gT \rvert + \lvert \gC \setminus \gT \rvert + \lvert \gT \setminus \gC \rvert
}\,.
\end{equation}
To characterize each sample's status within the cross-domain relationship graph, we introduce a structural score $\alpha$. This score is defined as the ratio of the expected connections to the noise-free set $\gT$, divided by the expected connections to the noisy set $\gN$. In formula, 
\begin{equation}\label{}
\alpha := \frac{
\sum_{k \in \gN(j) \cap \gT} w_{jk}
}{
\sum_{k \in \gN(j)} w_{jk}
}\,,
\end{equation}
where $w_{j,k}$ is for the edge weight of $\gW$. A smaller $\alpha_j$ indicates that the sample $j$ has more connections leading to unconfident samples, while a larger $\alpha_j$ suggests that the sample $j$ has more connections to the confident clusters.

Since the problem can be summarized into the $\ell_2$-norm flow diffusion problem.
In our cross-domain adaptation scenario, since the common knowledge is shared between domains, we have $a_0 \geq 1/2$ and $a_1 \geq 1/2$, which satisfy the condition that:
\begin{equation}
a_0 \geq 1 - \left( 
\sqrt{(\frac{\sum_{k \in \gN(j) \cap \gT} w_{jk}}{\alpha} + 2a_1-1)a_1} - a_1
\right)
\alpha + o_k(1) \,.
\end{equation}
Then we have the following Theorem, 
\begin{theorem}\label{eq:theorem-noise}
With $a_0$ and $a_1$ have lower bounded, we have the lower bound of $ACC\left( \gC\right)$ for each target data sample that: 
\begin{align}
&ACC\left( \gC\right) \notag \\
&\geq \left[ 1 + \frac{\left(1-a_1\right)}{2} + \frac{\left(1-a_0\right)}{2\alpha} + \frac{\left(1-a_0\right)^2}{2a_1\alpha^2}
\right]^{-1} - o_k\left(1\right) \,,
\end{align}
where $o_k(1)$ indicates that the term is a constant.
\end{theorem}

From the Theorem~\ref{eq:theorem-noise}, we can make the following observation.  Firstly, the quality of domain adaptation dynamics simulation is contingent upon the noise level. When the noise level in the target domain increases~(\ie, $\alpha$ increases), the quality of the obtained low-noise set $ACC(\gC)$ could decrease. Correspondingly, the higher values of $a_0$ and $a_1$ represent a better $ACC(\gC)$. More significantly, the quality of the low-noise set is guaranteed. When $\alpha$ is a constant~(\ie, $\alpha=\Omega_n(1)$), the $ACC(\gC)$ is lower-bounded by a constant, even if the target accuracy is very low~($a_0 \rightarrow 1/2$). This demonstrates the robustness of our method in identifying a reliable target set under noisy conditions. In conclusion, our graph flow diffusion method effectively leverages a non-parametric approach to model domain transfer dynamics, identifying a set with better connectivity throughout the process, and maintaining a superior confident set in the noisy target domain.

\paratitle{Deep Understanding of Graph Flow Diffusion.} 
Our flow diffusion approach is non-trivial for cross-domain hash retrieval. Beyond being the first to combine graph diffusion with cross-domain problems, we leverage diffusion to simulate and stabilize the adaptation process. This requires careful modeling of cross-domain relationships and dynamics rather than simple application of existing techniques.

\subsection{Discriminative Hash Learning}\label{sec::hash-learning}

In Section~\ref{sec:diffusion}, we employ graph diffusion for domain-adaptive dynamics simulation and identify \textit{early adopter} data with lower noise levels. In this section, we leverage these low-noise data to conduct cross-domain discriminative learning. This process is conducted in the source domain and the target domain.

\paratitle{Source Domain Hashing Learning.}
For source domain samples $\gD^s$, we directly use their true labels as supervision. These labels inherently represent explicit semantic information, effectively guiding the learning of discriminative hash codes. Specifically, an MLP $\phi(\cdot)$ is employed to map the latent variables to the Hamming space:
\begin{equation}
\vb=sign(\phi(\vf)) ~ \in \{-1, 1\}^L \,.
\end{equation}
The labels are projected into the Hamming space by,
\begin{equation}
\vz_{y_i} = sign(\phi(y_i)) ~ \in \{-1, 1\}^L \,.
\end{equation}
The label information and cross-entropy loss are then used for optimization:
\begin{equation}\label{eq:source-train}
\gL_s = - \sum_{i=1}^{\gN_s} \log\left(  \frac{
\exp\left( \vz_{y_i}^T \vb_i \right)
}{
\sum^{C}_{c=1} \exp \left( \vz_c^T \vb_i \right)
} \right)\,.
\end{equation}
As $sign(\cdot)$ is non-differentiable at zero and has zero gradients for all other inputs, $tanh(\cdot)$ is used to approximate $sign(\cdot)$ during the training process.

After sufficient training, it is possible to map from the label space to the Hamming space. Discriminative encodings are obtained by enforcing label embeddings to be discriminative and maximizing inter-class distances. To this end, the margin loss penalizes positively correlated label embedding pairs. 

\paratitle{Target Domain Hashing Learning.}
The labels are unavailable for the target domain. Therefore, we turn to pseudo-labels based on previously constructed cross-domain relation graphs.
Specifically, we use the reliable cluster $\gC$ extracted in Section~\ref{sec:diffusion}.
Samples in $\gC$ are composed of high-flow nodes from the cross-domain graph diffusion, representing well-connected regions in the graph. 
The lower noise set $\gC$ contains data samples from both the source and target domains.
Subsequently, we conduct intra-cluster discriminative learning via pseudo-label learning within the target confidence set $\gC$. The target domain pseudo-label $\hat{y}^t_j$ is generated by, 
\begin{equation}\label{eq:pseudo-label}
\hat{y}^t_j = \arg\max \vb^t_j \vz_c \,, 
\end{equation}
where $\vb^t_j$ is the hash code of image input $\vx^t_j$, and $\vz_c$ is the prediction of the category $c$. 

Then, we employ consistency learning for source and target samples within the same mini-batch. Specifically, let $\Pi(j)$ be the set of source samples holding the same label as $\vx^t_j$. The consistency learning object is written as,
\begin{equation}\label{eq:target-train}
\gL_t = - ~ \E_{j\in\gB^t} \frac{1}{\lvert\Pi(j)\rvert} 
\sum_{k\in\Pi(j)} \log\left(  \frac{
\exp\left( \vb_i^T \vb_k \right)
}{
\sum_{i\in\gB^s} \exp \left( {\vb_i^s}^T \vb_k^s \right)
} \right)\,,
\end{equation}
where $\mathcal{B}^t$ represents a mini-batch sampled from the target domain.
We jointly supervise the hash model by the labeled source domain data as well as the pseudo-labeled target domain samples. 
The overall discriminative learning objective is optimized for both source and target domains~(Eq.~\ref{eq:source-train} and Eq.~\ref{eq:target-train}), \ie,
\begin{equation}\label{eq:all-train}
\gL_{dis} = \gL_{s} + \gL_{t}\,.
\end{equation}
Through discriminative hash learning, we can leverage label and pseudo-label information to enable hash codes to effectively reconstruct the high-level semantics of images, thereby achieving effective image retrieval. Furthermore, discriminative hash learning, combined with knowledge from both domains, allows hash codes to encode cross-domain information effectively.

\subsection{Progressive Alignment with Cross-Domain Mixup}\label{sec::mixup}

\begin{figure}[t]
    \centering
    \includegraphics[width=0.95\linewidth]{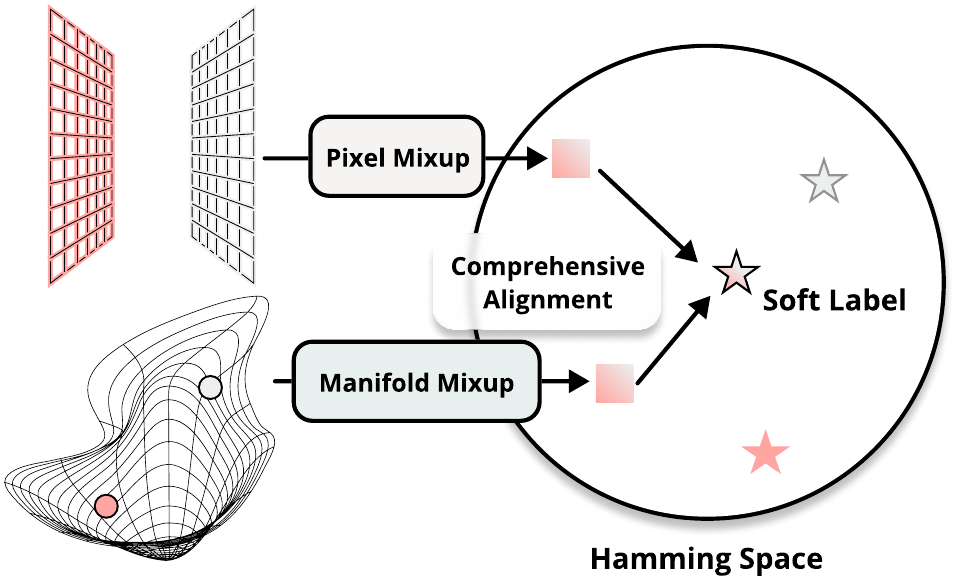}
    \caption{Hierarchical Mixup.
Low-level Mixup is for cross-domain pixel-level alignment, and high-level Mixup is to capture cross-domain semantic information. They contribute to the comprehensive adaptive hash codes under the different level domain shifts.}
    \label{fig:fig3}
\end{figure}

The techniques described in Section~\ref{sec::hash-learning} focus on generating discriminative hash codes. 
However, directly learning high-level hash codes for cross-domain data can be challenging to optimize.  To address this, we design a hierarchical Mixup operation on cross-domain Mixup pairs, facilitating progressive cross-domain alignment, as illustrated in Figure~\ref{fig:fig3}. We first outline the motivation behind our cross-domain Mixup mechanism. Next, we introduce Mixup pairs, explaining how to select the data for progressive domain alignment. Then, we detail our hierarchical Mixup operation. Finally, we present the optimization approach for this module.

\paratitle{Motivation.}
Domain discrepancies in real-world scenarios are often hierarchical and substantial. \fix{When the domain shift is too large, direct alignment methods may fail to bridge the gap effectively, necessitating a more gradual adaptation approach.} \fix{Additionally, domain shifts occur at multiple levels (e.g., pixel-level appearance and semantic-level concepts), requiring a comprehensive alignment strategy that can handle these different aspects simultaneously.} The hierarchical domain shits are in both semantic-level annd pixel-level. Existing methods frequently these hierarchical domain shifts, using direct domain alignment and thus complicating the learning process. We propose to address this challenge in two-fold. First, we introduce hierarchical Mixup to tackle multi-level domain discrepancies. Second, we employ progressive domain alignment along cross-domain random walk path on previously constructed cross-domain relation graph in Section~\ref{sec:diffusion}. This section aims to reduce the domain alignment difficulty.

\paratitle{Mixup Pairs.}
The Mixup operation typically requires pairwise data. 
In our \method{}, we leverage the cross-domain relationship graph and cross-domain flow diffusion to extract cross-domain relationships.

We sample the cross-domain pairwise data from the random walk across $\gD^s$ and $\gC$. The Mixup pair set $\gP$ is:
\begin{equation}\label{eq:random-walk}
\gP = \{\vx_0, \cdots ,\vx_k\}, ~\vx_0\in\gD^s, \vx_k\in\gC \,,
\end{equation}
where the random walk step $k$ is investigated in Sec.~\ref{sec:sensitivity} and set default to $5$. 
The transition probabilities are sampled from the distribution of edge weights as:
\begin{equation}
P\left(\vx_i\mid \vx_{i-1} \right) = \frac{
\vw^{(\vx_i, \vx_{i-1})}
}{
\sum_{\vx\in \gN(\vx_{i-1})} \vw^{(\vx_i, \vx)}
}\,.
\end{equation}
This represents the evolutionary process between two semantic clusters. By utilizing the neighbor pairs along the path, we construct a progressive cross-domain alignment process.

\begin{algorithm}[tb]
    \caption{Optimization of \method{}}
    \label{alg:algorithm}
    \begin{flushleft}
\textbf{Require}: Source domain $\gD^s$; Target domain $\gD^t$;\\
\textbf{Ensure}: Optimized model $\phi(\cdot)$; 
    \end{flushleft}
    \begin{algorithmic}[1] 
\STATE Warm up the model $\phi(\cdot)$ with Eq.~\ref{eq:source-train};
\FOR{$i$ = 1, 2, $\cdots$, $T$}
\STATE Conduct cross-domain graph diffusion process as Section~\ref{sec:diffusion};
\STATE Obtain the lower noise-level target set $\gC$;
\STATE Sample Mixup Paris $\gP$ by Eq.~\ref{eq:random-walk};
\FOR{each batch}
    \STATE Sample $\gB^s, \gB^t$ from $\gD^s, \gC$;
    \STATE Pseudo-labels generation using Eq.~\ref{eq:pseudo-label};
    \STATE Calculate $\gL_{dis}$ using Eq.~\ref{eq:all-train};
    \STATE Calculate $\gL_{mix}$ using Eq.~\ref{eq:mixup};
    \STATE Update parameters in the hash model $\phi(\cdot)$;
\ENDFOR
\ENDFOR
    \end{algorithmic}
\end{algorithm}

\paratitle{Hierarchical Mixup.}
Mixup pairs $\gP$ represent regions with significant semantic variations. Learning from these pairs is often challenging and difficult to optimize due to their sharp dissimilarities. 
Moreover, an effective hash encoding should consider high-level semantic information and low-level pixel information. Inspired by this, we propose a hierarchical Mixup operation~(Figure~\ref{fig:fig3}) tailored for hash encoding, which operates from both low and high levels.

\textit{For low-level Mixup}, we perform pixel-wise Mixup on images. In the cross-domain MNN graph established in Section~\ref{sec:diffusion}, edge weights represent the similarity between samples. Since the source domain samples are trustworthy, we can leverage the edge weights in the cross-domain relationship graph as the coefficients during this process:
\begin{equation}
\vx^m = \frac{1}{1+\vw^{(i,j)}}  \vx^i+ \frac{\vw^{(i,j)}}{1+\vw^{(i,j)}} \vx^j \quad (i,j)\in\gP \,,
\end{equation}
where sample $i$ is from the source domain and sample $j$ is from the target domain.
Simultaneously, to form effective soft supervision, we mix the labels using the same coefficient, \ie,
\begin{equation}\label{eq:mixup-label}
\vy^m = \frac{1}{1+\vw^{(i,j)}}  \vy^i+ \frac{\vw^{(i,j)}}{1+\vw^{(i,j)}} \hat{\vy}^j \quad (i,j)\in\gP \,,
\end{equation}
where $\vw^{(i,j)}$ denotes the edge weight of $i$ and $j$ in the cross-domain relationship graph. Then, we conduct pixel level Mixup for hash learning as:
\begin{equation}\label{eq:mixup-pixel}
\gL_{pixel} = - ~ \sum_{(i,j)\in\gP} \log\left(  \frac{
\exp\left( \vz_m^T \vb^m_{pixel}  \right)
}{
\sum^{C}_{c=1} \exp \left( \vz_c^T \vb^m_{pixel} \right)
} \right)\,.
\end{equation}
where $\vb^m_{pixel} = sign(\phi(F(\vx^m)))$.

Furthermore, in the manifold space~(high-level semantic space), the Mixup operation better represents the combination of different semantics in the Hamming space. For cross-domain samples, the combination can help the model gradually learn the target semantics. First, we perform the Mixup on the high-level feature as:
\begin{equation}
\vf^m = \frac{1}{1+\vw^{(i,j)}}  \vf^i+\frac{\vw^{(i,j)}}{1+\vw^{(i,j)}} \vf^j \,,
\end{equation}
where the $\vw^{(i,j)}$ is also the edge weight on the relationship graph, while the soft labels are generated as Eq.~\ref{eq:mixup-label}. We employ the manifold level Mixup as,
\begin{equation}\label{eq:mixup-manifold}
\gL_{manifold} = - ~ \sum_{(i,j)\in\gP} \log\left(  \frac{
\exp\left( \vz_m^T \vb^m_{manifold}  \right)
}{
\sum^{C}_{c=1} \exp \left( \vz_c^T \vb^m_{manifold} \right)
} \right)\,,
\end{equation}
where $\vb^m_{manifold} = sign(\phi(\vf^m))$.

Then, we perform the Mixup operation for progressive domain alignment on intra-cluster and inter-cluster cross-domain samples $\gP$, and in a progressive manner:
\begin{equation}\label{eq:mixup}
\gL_{mix} = \gL_{pixel} + \gL_{manifold}\,.
\end{equation}

We learn the intermediate transferred variables by optimizing $\gL_{mix}$, enabling progressive transfer learning. 

\paratitle{Deep Understanding of Hierarchical Mixup.} 
\fix{The hierarchical Mixup mechanism is non-trivial in cross-domain hash retrieval for its dual-granularity design. The pixel-level Mixup bridges low-level visual differences, while semantic-level Mixup enables gradual adaptation of high-level concepts. More importantly, by leveraging the graph structure from diffusion, the Mixup paths follow meaningful cross-domain transitions rather than arbitrary mixing.}
\fix{The Mixup and Diffusion modules work synergistically in three aspects: (1) Diffusion provides reliable target samples for meaningful Mixup operations, (2) the graph structure guides natural paths for progressive Mixup (Eq.~\ref{eq:random-walk}), and (3) while Diffusion handles global structure, Mixup creates local smooth transitions for effective hierarchical adaptation.}

\subsection{Overall Algorithm}

In this section, we introduce the overall algorithm of our proposed method. By designing a cross-domain diffusion process to simulate adaptation dynamics, we revisit domain adaptive hashing learning. 
This approach enables us to divide and conquer target domain data with varying noise levels, achieving robust target domain adaptation. 
Specifically, we construct relationship graphs for the source and target domains, and then leverage a graph diffusion process to identify \textit{early adaptater} data samples in the target domain. For these samples, we fully exploit the target domain data via discriminative learning based on pseudo-labels. Simultaneously, we design a progressive domain adaptation process by utilizing the Mixup operation to find cross-domain paths between data points and progressively transfer across these random-walk paths via Mixup. Our algorithm is summarized in Algorithm~\ref{alg:algorithm}.

\section{Experiment}
\label{sec::experiment}

\subsection{Experimental Settings}

\paratitle{Datasets.} We conducted experiments on benchmark datasets, primarily including three datasets:
\begin{itemize}
    \item \textit{Office-Home}~\cite{venkateswara2017deep}, containing images from four different domains: Artistic~(Ar), Clip Art~(Cl), Product~(Pr), and Real-World~(Re). We selected two different domains as the source and target domains. For comparison with previous works, we performed domain transfer image retrieval on $6$ transfer tasks. They are: Pr$\rightarrow$Re, Cl$\rightarrow$Re, Re$\rightarrow$Ar, Re$\rightarrow$Pr, Re$\rightarrow$Cl, Ar$\rightarrow$Re.
    \item \textit{Office-31}~\cite{saenko2010adapting}, containing 31 categories with over $4,000$ images from three domains: Amazon~(Am), Webcam~(We), and DSLR~(Ds). Office-31 is a standard domain adaptation dataset, including $6$ transfer tasks for image retrieval. They are: Am$\rightarrow$Ds, Am$\rightarrow$We, We$\rightarrow$Ds, Ds$\rightarrow$Am, We$\rightarrow$Am, Ds$\rightarrow$We.
    \item  \textit{Digits}, where we studied transfer learning on the MNIST~\cite{lecun1998gradient} and USPS~\cite{hull1994database} digit datasets, alternating them as source and target domains, resulting in two transfer tasks (MNIST$\rightarrow$USPS and USPS$\rightarrow$MNIST).
    \item \fix{\textit{DomainNet}~\cite{peng2019moment}, a comprehensive cross-domain dataset across six distinct visual domains, with about $600,000$ images. We focus on four domains: Real~(Re), Clipart~(Cl), Painting~(Pa), and Sketch~(Sk).}
\end{itemize}

\begin{table*}[t]
\centering
\tabcolsep=1.7pt
\caption{Cross-domain retrieval performance~(mAP$\%$) comparison on the Office-Home and Office-31 datasets. }
\resizebox{\textwidth}{!}{
\begin{tabular}{lrcccccccccccccc}
\toprule[1pt]
\multicolumn{2}{c}{\multirow{2}{*}{\bf{Methods}}} & \multicolumn{6}{c}{\bf{OFFICE-HOME}} && \multicolumn{6}{c} {\bf{OFFICE-31}} & 
\\
\cmidrule{3-8} \cmidrule{10-15}
&& Pr$\rightarrow$Re & Cl$\rightarrow$Re & Re$\rightarrow$Ar & Re$\rightarrow$Pr & Re$\rightarrow$Cl & Ar$\rightarrow$Re 
&
& Am$\rightarrow$Ds & Am$\rightarrow$We & We$\rightarrow$Ds & Ds$\rightarrow$Am & We$\rightarrow$Am & Ds$\rightarrow$We &\bf{Avg.} \\
\midrule 
ITQ & {\scriptsize TPAMI'13} & 
26.81 & 14.83 & 25.37 & 28.19 & 14.92 & 25.88 &&
29.55 & 28.53 & 58.00 & 26.83 & 25.09 & 58.89 & 30.24 \\
OCH & {\scriptsize TPAMI'19} & 
18.65 & 10.27 & 17.54 & 20.15 & 10.05 & 18.09 &&
24.86 & 22.49 & 51.03 & 22.45 & 20.79 & 53.64 & 24.17 \\
DSH & {\scriptsize TCYB'14} & 
8.49 & 5.47 & 9.67 & 8.26 & 5.28 & 9.69 &&
16.66 & 15.09 & 39.24 & 16.33 & 13.58 & 41.07 & 15.74 \\
SGH & {\scriptsize IJCAI'15} & 
24.51 & 13.62 & 22.53 & 25.73 & 13.51 & 22.93 &&
24.98 & 22.47 & 53.94 & 22.17 & 20.52 & 56.36 & 26.94 \\
GraphBit & {\scriptsize TPAMI'23} &
18.18 & 16.87 & 11.51 & 10.81 & 18.91 & 21.32 &&
24.48 & 23.12 & 22.09 & 53.82 & 21.34 & 51.43 & 24.49 \\
GTH-g & {\scriptsize TCSVT'21} &
20.00 & 10.99 & 18.28 & 21.95 & 11.68 & 19.05 && 23.08 & 21.20 & 49.38 & 19.52 & 17.41 & 50.14 & 23.56 \\
PWCF & {\scriptsize CVPR'20} &
34.03 & 24.22 & 28.95 & 34.44 & 18.42 & 34.57 && 39.78 & 34.86 & 67.94 & 35.12 & 35.01 & 72.91 & 38.35 \\
DHLing & {\scriptsize MM'21} &
48.47 & 30.81 & 38.68 & 45.24 & 25.15 & 43.30 && 41.96 & 45.10 & 75.23 & 42.89 & 41.74 & 79.91 & 46.54 \\
DAPH & {\scriptsize TNNLS'22} &
27.20 & 15.29 & 27.35 & 28.19 & 15.29 & 26.37 && 32.80 & 28.66 & 60.71 & 28.66 & 27.59 & 64.11 & 31.85 \\
PEACE & {\scriptsize TIP'23} & 
53.04 & 38.72 & 42.68 & 54.39 & 28.36 & 45.97 && 46.69 & 48.89 & 78.82 & 46.91 & 46.95 & 83.18 & 51.22 \\
DANCE & {\scriptsize WWW'23} & 
53.73 & 39.03 & 43.54 & 55.14 & 28.87 & 44.53 && 
44.78 & 47.66 & 78.39 & 46.68 & 48.61 & 84.75 & 51.31 \\
IDEA & {\scriptsize NeurIPS'23} &
59.18 & 45.71 & 49.64 & 61.84 & 32.77 & 51.19 && 48.70 & 54.43 & 84.97 & 53.53 & 53.71 & 88.69 & 57.03 \\
\midrule
\rowcolor{LightCyan} \method{} & {\scriptsize Ours } & \bf 63.94 & \bf 49.24 & \bf 54.35 & \bf 64.29 & \bf 41.39 & \bf 54.14 &&
\bf 50.27 & \bf 59.32 & \bf 85.26 & \bf 56.04  & \bf 56.35 & \bf 88.90 &  \bf 60.29
\\
\bottomrule[1pt]
\end{tabular}
}
\label{tab:maintab1}
\end{table*}

\begin{figure*}[!ht]
    \centering
    \includegraphics[width=\textwidth]{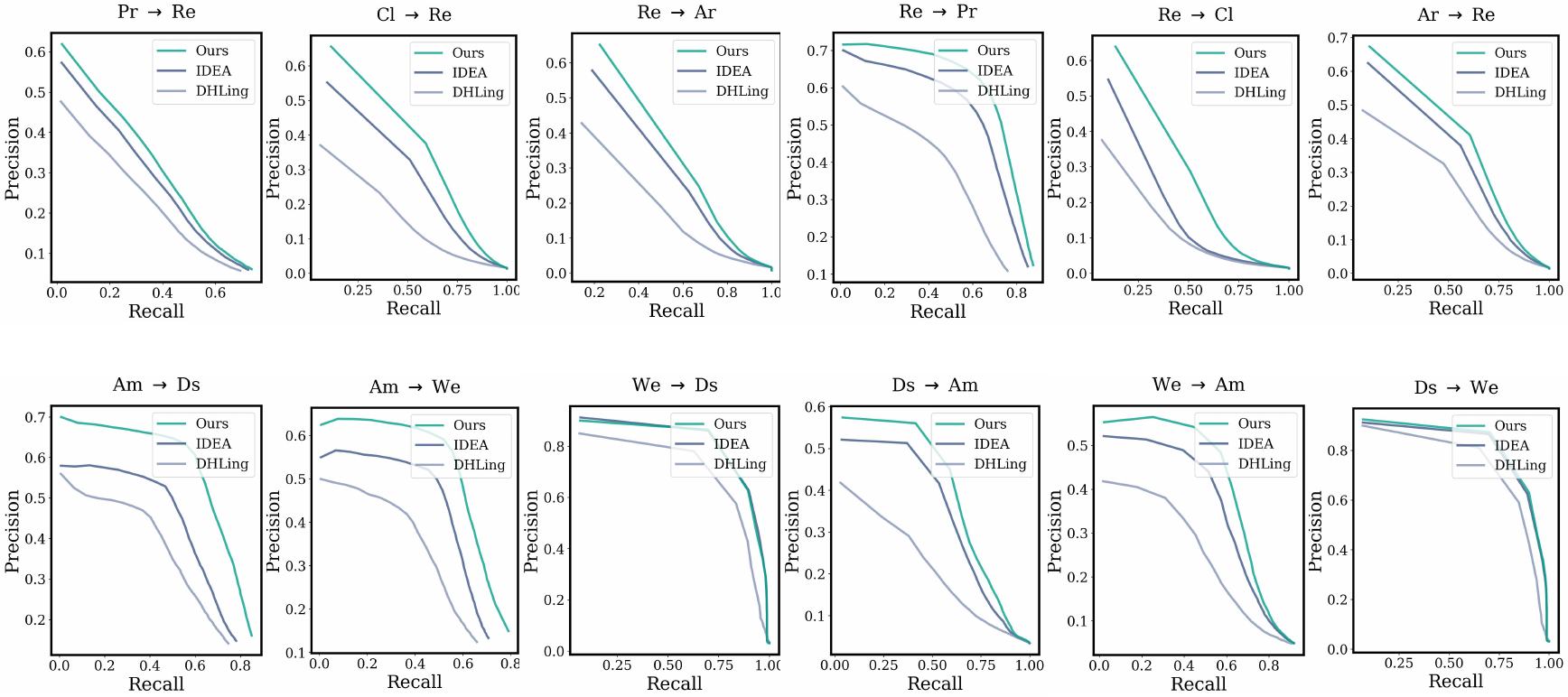}
    \caption{Precision-recall curves with $64$ bits hash code on Office-Home and Office-31 datasets.}
    \label{fig:exp-curve-pr}
\end{figure*}

\paratitle{Compared Baselines.} We compare \method{} with the state-of-the-art domain adaptive hashing methods, including unsupervised hashing methods and adaptive hashing method. The results of these methods are consistent with those reported in their original papers. The baseline methods are detailed as follows:
\begin{itemize}
    \item ITQ~\cite{gong2012iterative}: A simple yet effective alternating minimization algorithm that can incorporate both supervised and unsupervised learning processes.
    \item DSH~\cite{jin2013density}: A variant of LSH that generates multiple views of data points using random projections for metric learning.
    \item SGH~\cite{jiang2015scalable}: Aims to compress high-dimensional data in a bit-wise manner, effective for large-scale semantic similarity learning.
    \item GraphBit~\cite{wang2022learning}: Mines bit-wise interactions in the continuous space, significantly reducing the expensive search cost caused by training difficulties in reinforcement learning.
    \item GTH-g~\cite{zhang2019optimal}: Selects the optimal hash mapping for target data from source data based on the maximum likelihood estimation principle.
    \item PWCF~\cite{huang2020probability}: Learns discriminative hash codes via a Bayesian model and infers similarity structure using histogram features.
    \item DHLing~\cite{mm21-adahash-xia}: Optimizes hash codes within a single domain via learned clustering, then utilizes a memory bank to reduce domain shift.
    \item DAPH~\cite{huang2021domain}: Reduces domain discrepancy via domain-invariant feature projection.
    \item PEACE~\cite{wang2023toward}: Learn target data semantics using pseudo-labeling techniques, then minimize domain transfer in implicit and explicit manners.
    \item DANCE~\cite{wang2023dance}: Dual-level hash learning, which measures prototypes of high-level features across domains and is optimized with constrastive learning.
    \item IDEA~\cite{wang2024idea}: Decomposes each visual vector into causal features representing label information and non-causal features, generates hash codes using causal features.
\end{itemize}

\begin{table*}[t]
\centering
\tabcolsep=5pt
\caption{Cross-domain retrieval performance~(mAP$\%$) comparison on the MNIST and USPS datasets.}
\resizebox{\textwidth}{!}{
\begin{tabular}{lrcccccccccccccc}
\toprule[1pt]
\multicolumn{2}{c}{\multirow{2}{*}{\bf{Methods}}} & \multicolumn{6}{c}{\bf{MNIST $\rightarrow$ USPS}} && \multicolumn{6}{c} {\bf{USPS $\rightarrow$ MNIST}} & 
\\
\cmidrule{3-8} \cmidrule{10-15}
&& 16 & 32 & 48 & 64 & 96 & 128 && 
16 & 32 & 48 & 64 & 96 & 128 &\bf{Avg.} \\
\midrule 
ITQ & {\scriptsize TPAMI'13} & 
13.05 & 15.57 & 18.54 & 20.12 & 23.12 & 23.89 &&
13.69 & 17.51 & 20.40 & 20.30 & 22.79 & 24.59 & 19.46 \\
OCH & {\scriptsize TPAMI'19} & 
13.73 & 17.22 & 19.59 & 20.18 & 20.66 & 23.34 && 
15.51 & 17.75 & 18.97 & 21.50 & 21.27 & 23.68 & 19.45 \\
DSH & {\scriptsize TCYB'14} & 
20.60 & 22.21 & 23.68 & 24.28 & 25.73 & 26.50 &&
19.54 & 21.22 & 22.89 & 23.79 & 25.91 & 26.46 & 23.57 \\
SGH & {\scriptsize IJCAI'15} & 
14.24 & 16.69 & 18.72 & 19.70 & 21.00 & 21.95 &&
13.26 & 17.71 & 18.22 & 19.01 & 21.69 & 22.09 & 18.69 \\
GraphBit & {\scriptsize TPAMI'23} &
13.92 & 17.86 & 20.17 & 20.82 & 21.32 & 23.19 &&
15.16 & 16.82 & 17.87 & 19.85 & 20.10 & 22.54 & 19.13 \\
GTH-g & {\scriptsize TCSVT'21} &
20.45 & 17.64 & 16.60 & 17.25 & 17.26 & 17.06 && 
15.17 & 14.07 & 15.02 & 15.01 & 14.80 & 17.34 & 16.47 \\
PWCF & {\scriptsize CVPR'20} &
47.47 & 51.99 & 51.44 & 51.75 & 50.89 & 59.35 && 47.14 & 50.86 & 52.06 & 52.18 & 57.14 & 58.96 & 52.60 \\
DHLing & {\scriptsize MM'21} &
49.24 & 54.90 & 56.30 & 58.28 & 58.80 & 59.14 && 50.14 & 51.35 & 53.67 & 58.65 & 58.42 & 59.17 & 55.67 \\
DAPH & {\scriptsize TNNLS'22} &
25.13 & 27.10 & 26.10 & 28.51 & 30.53 & 30.70 && 26.60 & 26.43 & 27.27 & 27.99 & 30.19 & 31.40 & 28.16 \\
PEACE & {\scriptsize TIP'23} & 
52.87 & 59.72 & 60.69 & 62.84 & 65.13 & 68.16 && 53.97 & 54.82 & 58.69 & 60.91 & 62.65 & 65.70 & 60.51 \\
DANCE & {\scriptsize WWW'23} & 
53.18 & 57.98 & 61.23 & 63.15 & 65.92 & 68.87 && 54.31 & 55.64 & 57.26 & 61.49 & 63.43 & 66.23 & 60.72  \\
IDEA & {\scriptsize NeurIPS'23} &
58.89 & 64.48 & 65.72 & 67.48 & 70.24 & 74.34 && 60.99 & 61.47 & 65.45 & 67.97 & 69.72 & 72.31 & 66.59 \\
\midrule
\rowcolor{LightCyan} \method{} & {\scriptsize Ours } & \bf 60.56  & \bf 66.05  & \bf 66.23  & \bf 67.98 & \bf 73.02 & \bf 75.12 &&
 \bf 63.28 &  \bf 64.94  &  \bf 67.44  & \bf 70.19  &  \bf 72.87 & \bf 74.62  &  \bf 68.53
\\
\bottomrule[1pt]
\end{tabular}
}
\label{tab:maintab2}
\end{table*}

\begin{figure*}[t]
    \centering
    \includegraphics[width=\textwidth]{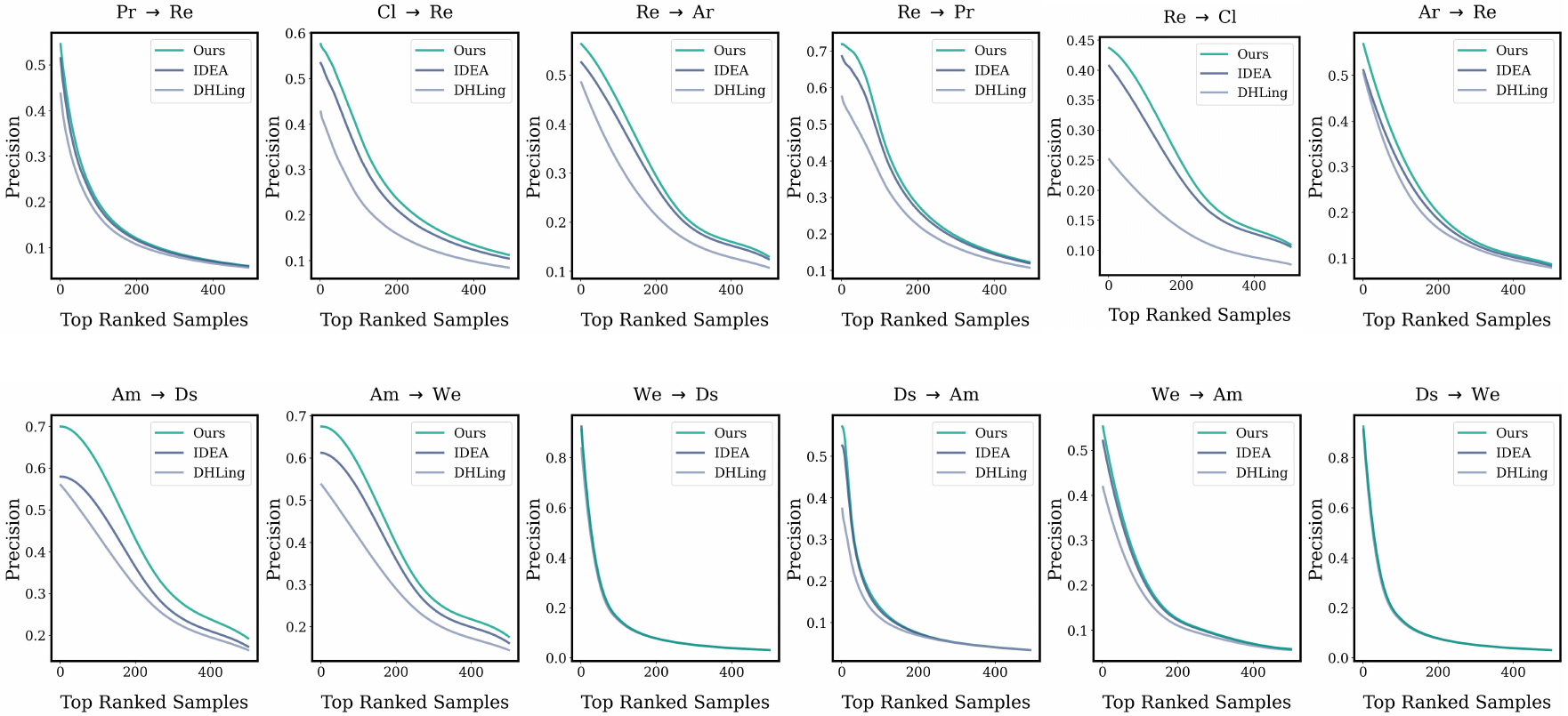}
    \caption{ Top-N precision curves with $64$ bits hash code on Office-Home and Office31 datasets.}
    \label{fig:curve-topn-pr}
\end{figure*}
{
\begin{table*}[tb]
\centering
\tabcolsep=5pt
\caption{\fix{Ablation studies on the Office-Home with $64$ bit hash code.}}
\begin{tabular}{lcccccccccccccc}
\toprule[1pt]
Variants
&& PL & CD & PMix & MMix & MT && Pr$\rightarrow$Re & Cl$\rightarrow$Re & Re$\rightarrow$Ar & Re$\rightarrow$Pr & Re$\rightarrow$Cl & Ar$\rightarrow$Re & \bf{Avg.}
\\
\midrule 
\method{}~\textit{V1} && \cmark &&&& && 56.18 & 43.66 & 49.54 & 59.55 & 35.80 & 49.17 & 48.98 \\ 
\method{}~\textit{V2} && \cmark & \cmark &&& && 
61.18 & 47.69 & 53.03 & 63.03 & 38.48 & 52.47 & 52.65 \\ 
\method{}~\textit{V3} && \cmark &  & \cmark& &&& 
57.60 & 45.22 & 50.68 & 61.04 & 37.37 & 50.90 & 50.47 \\
\method{}~\textit{V4} && \cmark & \cmark & \cmark & &&&  62.29 & 48.65 & 53.27 & 63.77 & 40.53 & 52.90 & 53.57 \\ 
\method{}~\textit{V5} && \cmark & && \cmark &&& 
58.54 & 45.84 & 51.92 & 61.93 & 38.17 & 53.55 & 52.13 \\ 
\method{}~\textit{V6} && \cmark & \cmark && \cmark &&& 
63.01 & 48.78 & 53.38 & 63.67 & 40.71 & 53.55 & 53.85 \\ 
\method{}~\textit{V7} && \cmark &  &&  & \cmark && 
57.44 & 45.07 & 51.32 & 61.24 & 37.92 & 51.38 & 50.73 \\ 
\midrule 
\method{}~(Full Model) && \cmark & \cmark & \cmark & \cmark &&&  \textbf{63.94} & \textbf{49.24} & \textbf{54.35} & \textbf{64.29} & \textbf{41.39} & \textbf{54.14} & \textbf{54.56}
\\
\bottomrule[1pt]
\end{tabular}
\label{tab:abl}
\end{table*}
}
\begin{figure*}[t]
    \centering
    \includegraphics[width=\textwidth]{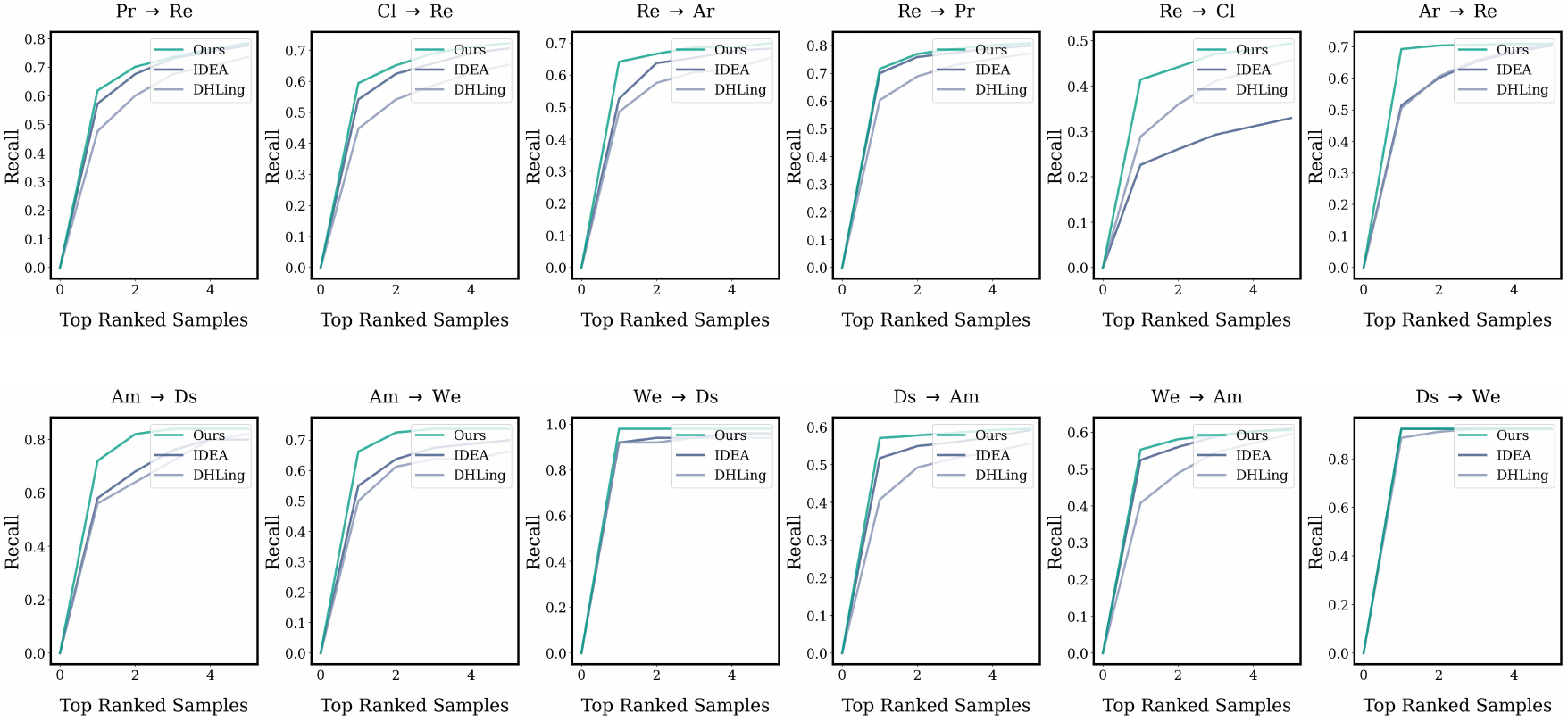}
    \caption{ Top-N recall curves with $64$ bits hash code on Office-Home and Office31 datasets.}
    \label{fig:curve-topn-recall}
\end{figure*}

\begin{figure*}[!ht]
    \centering
    \includegraphics[width=\linewidth]{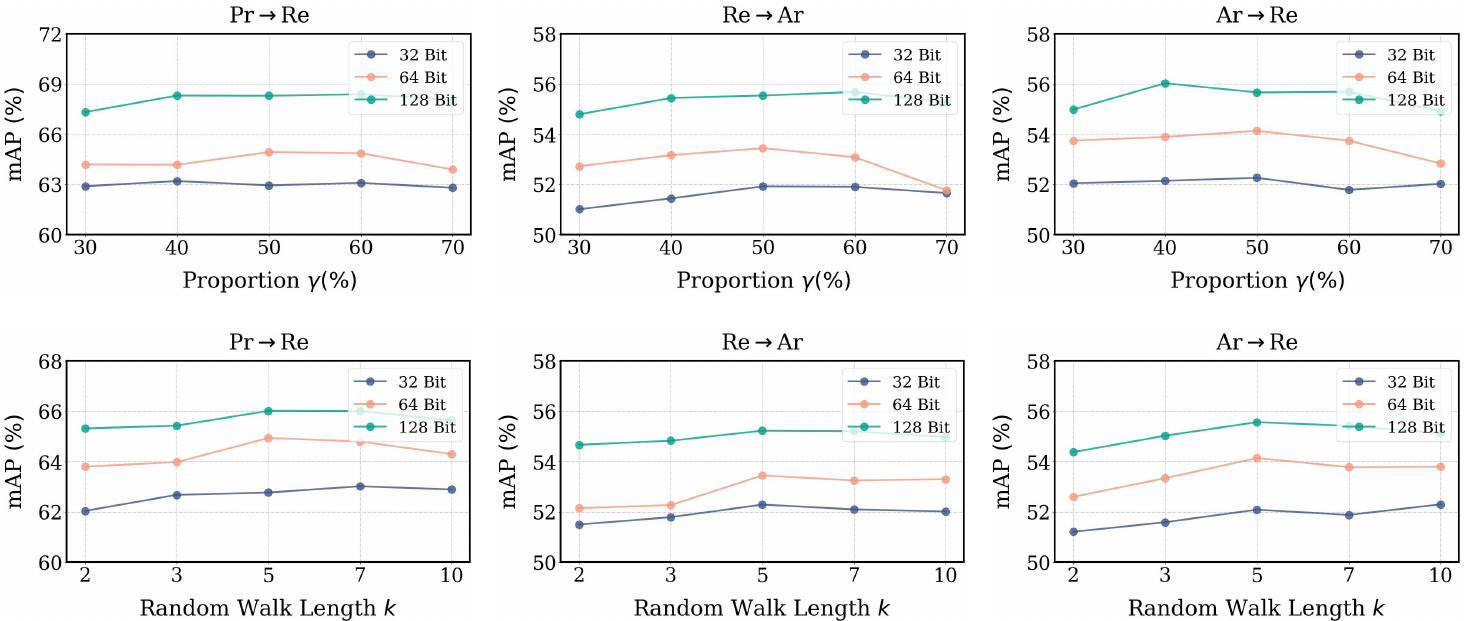}
    \caption{Sensitivity analysis on $\gamma$ and $k$.}
    \label{fig:abl-sensitive}
\end{figure*}

\paratitle{Implementation Details.} To ensure a fair comparison, the model config is set following previous methods~\cite{mm21-adahash-xia,wang2023toward,wang2024idea}. \fix{Following previous works, we use VGG-F as the visual feature encoder for consistent comparison.} Specifically, our hash encoder employs a two-layer MLP model $\phi(\cdot)$ to project the visual feature into the Hamming space. Additionally, we employ the $tanh(\cdot)$ activation function instead of $sign(\cdot)$. Our implementation is based on PyTorch, utilizing the Adam optimizer with a learning rate of $0.001$ and a batch size of $32$. We set proportion $\gamma$ to $0.5$, random walk length $k$ to $5$, according to the hyper-parameter studies in Sec.~\ref{sec:sensitivity}.
All experiments were conducted on an NVIDIA RTX 4090 GPU. 

\paratitle{Evaluation Metrics.} We employ three standard metrics to evaluate the retrieval performance of our \method{}: mean Average Precision~(mAP), precision-recall curve, Top-N accuracy curve and Top-N recall curve. The mAPs represent the overall retrieval performance, the precision-recall curves assess the comprehensive performance of the method, the Top-N accuracy curves and the Top-N recall curves illustrats the performance under different retrieval quantities. 

\subsection{Performance Comparison}

To comprehensively validate the effectiveness of our proposed \method{}, we conducted both quantitative and qualitative comparisons with current state-of-the-art approaches. 

\paratitle{Quantitative Performance.}
Our quantitative experiments encompassed two main perspectives. First, we evaluated performance on the Office-Home and Office-31 cross-domain tasks using a fixed $64$-bit hash code length, as presented in Table~\ref{tab:maintab1}. Second, we examined cross-domain performance on USPS and MNIST datasets across varying hash code lengths, with results shown in Table~\ref{tab:maintab2}. These comprehensive evaluations provide a robust assessment of our method's efficacy in diverse cross-domain retrieval scenarios.

From Table~\ref{tab:maintab1} and \ref{tab:maintab2}, we observed that our proposed method significantly outperformed existing state-of-the-art approaches in cross-domain retrieval, achieving an average improvement of over $3\%$. 
Notably, substantial gains were observed in challenging subtasks where previous methods struggled, such as Re$\rightarrow$Cl~($8.6\%$ improvement) and Pr$\rightarrow$Re~($4.8\%$ improvement) in Office-Home, and Am$\rightarrow$We~($4.9\%$ improvement) and We$\rightarrow$Am ($2.6\%$ improvement) in Office-31. \fix{Similar improvements were also observed in Table~\ref{tab:domainnet} on the large-scale DomainNet dataset, where \method{} achieved an average mAP of $61.26\%$, surpassing IDEA and PEACE across five challenging transfer tasks.} \fix{The relatively lower performance of early methods like DAPH~\cite{huang2021domain} can be attributed to their simpler domain adaptation strategies, which may not be as effective in handling complex domain discrepancies compared to recent advanced techniques.} 
These improvements can be attributed to two key factors. On the one hand, our \method{} is nosie-rubst in adaptive retrieval, particularly in cases of significant domain discrepancies. \method{}'s carefully designed progressive domain alignment approach, which adeptly handles hierarchical domain differences for stable target domain adaptation. Furthermore, our method consistently maintained its superiority across various hash code lengths, demonstrating effective utilization of hash code capacity for retrieval. Importantly, we observed more improvements at lower bit lengths~(\ie, $16$, $32$) compared to existing methods, which further shows \method{}'s efficacy in handling domain discrepancies and achieving superior retrieval performance.

{
\begin{table}[t]
\centering
\tabcolsep=3pt
\caption{\fix{Retrieval time cost~(ms) varies with code length.}}
\begin{tabular}{lcccccc}
\toprule[1pt]

& $16$ Bit & $32$ Bit & $48$ Bit & $64$ Bit & $96$ Bit & $128$ Bit
\\
\midrule 
\textit{Hash} Code & 16.68 & 18.04 & 19.42 & 19.92 & 21.81 & 22.20 \\ 
\textit{Dense} Vector & 441.4 & 491.0 & 543.0 & 602.3 & 657.7 & 696.6 \\
Speed Up & 26.46$\times$ & 27.21$\times$ & 27.96$\times$ &30.23$\times$ &30.15$\times$ &31.38$\times$
\\
\bottomrule[1pt]
\end{tabular}
\label{tab:time}
\end{table}
}
\begin{table}[t]
\centering
\tabcolsep=5pt
\caption{\fix{Cross-domain retrieval performance~(mAP$\%$) comparison on the DomainNet dataset.}}
\begin{tabular}{lcccccc}
\toprule[1pt]
{\bf{Methods}} & Re$\rightarrow$Sk & Sk$\rightarrow$Pa & Pa$\rightarrow$Cl & Cl$\rightarrow$Re & Cl$\rightarrow$Pa & \bf{Avg.} \\
\midrule
PEACE & 51.92 & 54.13 & 52.73 & 73.02 & 50.68 & 56.50 \\
IDEA & 54.99 & 55.49 & 54.91 & 74.11 & 54.10 & 58.72 \\
\rowcolor{LightCyan} \method{} & \bf 56.72 & \bf 58.48 & \bf 58.45 & \bf 76.90 & \bf 55.77 & \bf 61.26 \\
\bottomrule[1pt]
\end{tabular}
\label{tab:domainnet}
\end{table}
{
\begin{table}[t]
\centering
\tabcolsep=3pt
\caption{\fix{Performance comparison of different backbones on various datasets.}}
\begin{tabular}{lcccccc}
\toprule

& \multicolumn{2}{c}{Office-Home} & \multicolumn{2}{c}{Office31} & \multirow{2}{*}{Avg.} \\
\cmidrule(lr){2-3} \cmidrule(lr){4-5}
Method & Cl$\rightarrow$Re & Re$\rightarrow$Ar & Am$\rightarrow$Ds & Am$\rightarrow$We & \\
\midrule
PEACE (VGG-F) & 38.72 & 42.68 & 46.69 & 48.89 & 44.25 \\
IDEA (VGG-F) & 45.71 & 49.64 & 48.70 & 54.43 & 49.62 \\
COUPLE (VGG-F) & 49.24 & 54.35 & 50.27 & 59.32 & 53.30 \\
\midrule
PEACE (ResNet-34) & 46.32 & 48.91 & 51.75 & 52.73 & 49.93 \\
IDEA (ResNet-34) & 48.77 & 51.06 & 53.88 & 56.80 & 52.63 \\
COUPLE (ResNet-34) & 52.79 & 56.39 & 55.90 & 61.27 & 56.59 \\
\midrule
PEACE (ViT-Base) & 53.83 & 56.11 & 56.05 & 60.25 & 56.56 \\
IDEA (ViT-Base) & 58.56 & 64.39 & 63.27 & 66.30 & 63.13 \\
COUPLE (ViT-Base) & 62.70 & 66.41 & 66.12 & 68.33 & 65.89 \\
\bottomrule
\end{tabular}
\label{tab:backbone_comparison}
\end{table}
}
\begin{figure}[t]
    \centering
    \includegraphics[width=\linewidth]{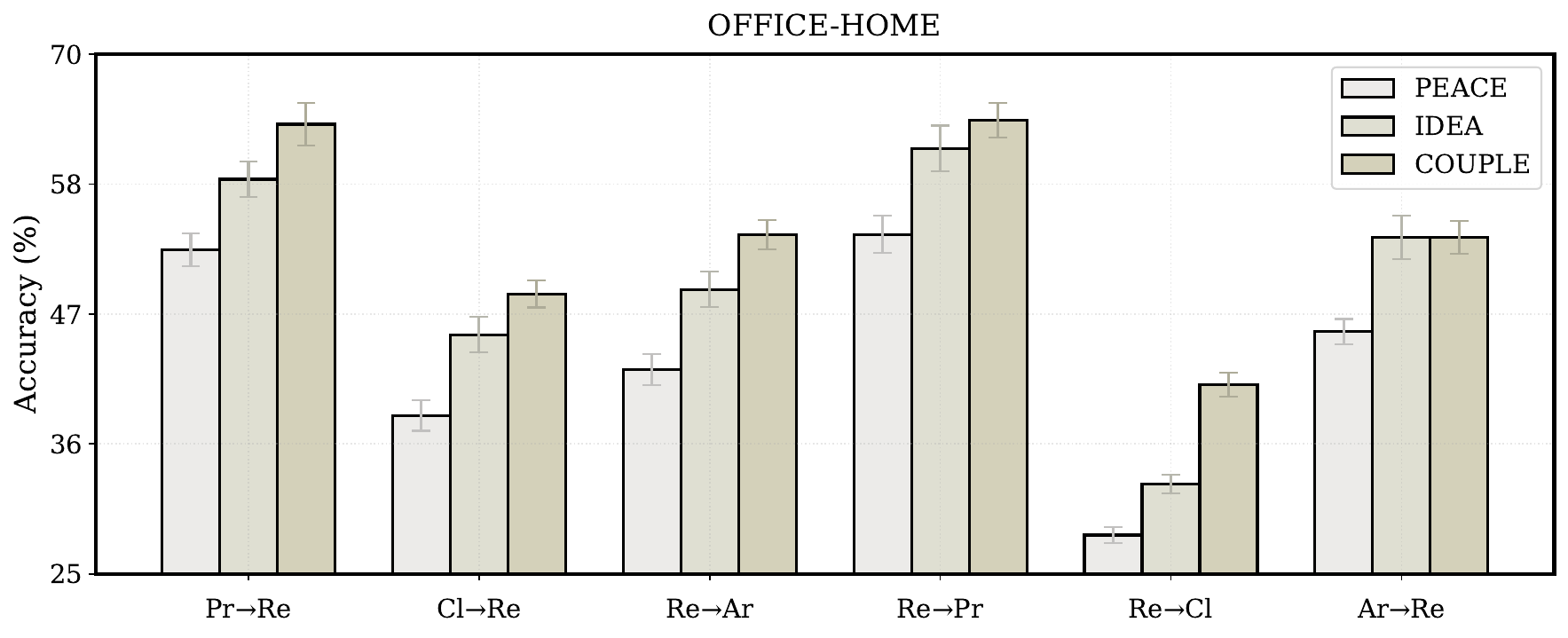}
    \caption{\fix{Stability analysis on Office-Home dataset.}}
    \label{fig:stability}
\end{figure}

\paratitle{Qualitative Analysis.} To gain deeper insights into the effectiveness of our proposed \method{}, we compared different methods using precision-recall curves, Top-N precision curves and Top-N recall curves, as illustrated in Figure~\ref{fig:exp-curve-pr}, ~\ref{fig:curve-topn-pr} and \ref{fig:curve-topn-recall}. \fix{The precision-recall curves evaluate the trade-off between precision and recall across all retrieval thresholds, while the Top-N precision curves focus on the retrieval accuracy for the most relevant results that users first encounter.} 
It is evident that our method consistently outperformed the baselines in cross-domain retrieval tasks, exhibiting larger areas under the curves compared to the competing methods. Notably, \method{} achieves higher precision at lower recall levels, which is crucial in real-world scenarios like browsing mobile-phone galleries and online shopping, where user attention is limited. 
Furthermore, our method yields superior precision and recall rates for the same sample sizes. This demonstrates its ability to maximize the potential of limited hash codes, providing accurate retrieval even with smaller sample sets, thus highlighting its practicality. The qualitative analysis, encompassing comparisons from various perspectives, emphasizes the potential of \method{} as a promising solution across diverse application scenarios.

\subsection{Ablation Study}
\label{sec:abl}

To valid the submodules, we introduced the following variants of \method{}: 
\textit{V1,} which only uses target domain pseudo-labeling;
\textit{V2,} which adds cross-domain diffusion based on \textit{V1};
\textit{V3,} which uses pixel-level Mixup and pseudo-labeling;
\textit{V4,} which combines pixel-level Mixup and cross-domain diffusion;
\textit{V5,} which applies manifold-level Mixup based on \textit{V1}; 
\textit{V6,} which combines manifold-level Mixup and cross-domain diffusion;
\fix{\textit{V7,} which employs mean teacher for pseudo-label refinement;}
The full model of \method{} utilizes all components.

We conducted ablation experiments using different variants, as shown in Table~\ref{tab:abl}, using the Office-Home dataset. The results lead to the following conclusions:
\begin{itemize}
    \item The \method{}~(full model) is at the best performance compared to its variants, demonstrating the importance of each component.
    \item Comparing \textit{V2} to \textit{V1} reveals the effect of cross-domain diffusion for select target domain data at different noise levels. This observation also exists when comparing \textit{V4} to \textit{V3} and \textit{V6} to \textit{V5}.
    \item Comparing \textit{V4} and \textit{V2} shows the role of pixel-level Mixup. For different domains, the Mixup operation can effectively achieve progressive domain alignment. Similar observations can be made when comparing \textit{V3} and \textit{V1}.
    \item The manifold-level mixup eliminates semantic-level domain differences, enabling effective domain transfer. These observations exist when comparing \textit{V6} to \textit{V2} and \textit{V5} and \textit{V1}.
    \item Manifold-level Mixup has a higher contribution, as seen when comparing \textit{V6} to \textit{V4} or \textit{V5} to \textit{V2}. 
    \fix{\item The mean teacher variant~(\textit{V7}) shows moderate improvement over the basic pseudo-labeling~(\textit{V1}). This demonstrates that mean teacher is effective for noise reduction in pseudo-labels, though not as powerful as our proposed hierarchical domain alignment strategy.}
    \item However, the full model, combining all approaches, demonstrates superior performance, indicating the hierarchical nature of domain differences.
\end{itemize}

\begin{figure*}[t]
    \centering
    \includegraphics[width=\linewidth]{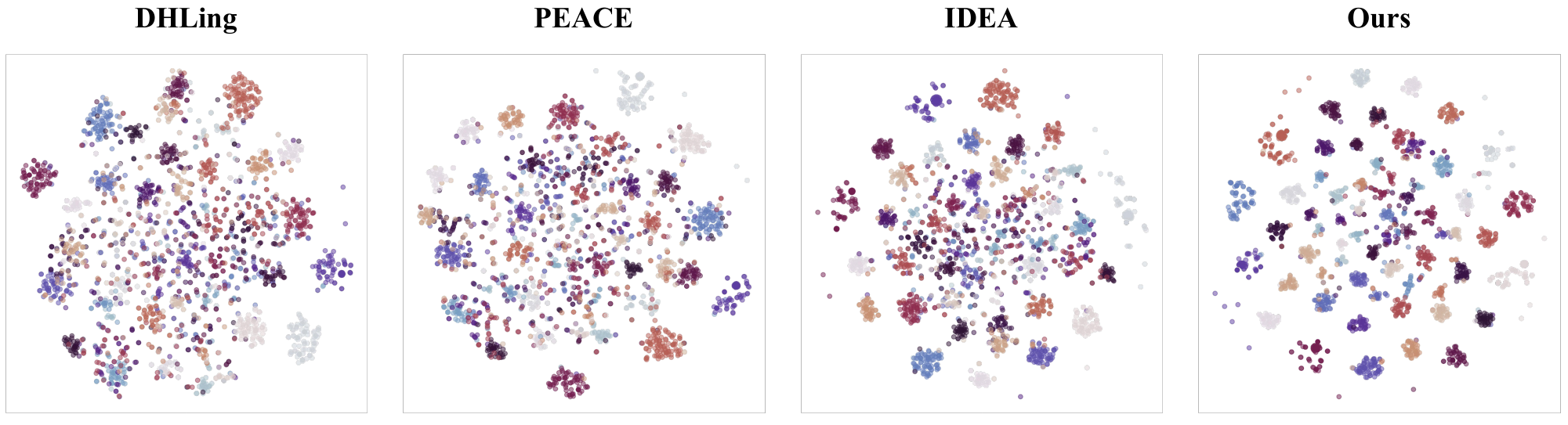}
    \caption{The t-SNE visualization of 64-bit hash codes on Office-Home dataset~(Ar $\rightarrow$ Re).}
    \label{fig:tsne}
\end{figure*}

\begin{figure}[t]
    \centering
    \includegraphics[width=\linewidth]{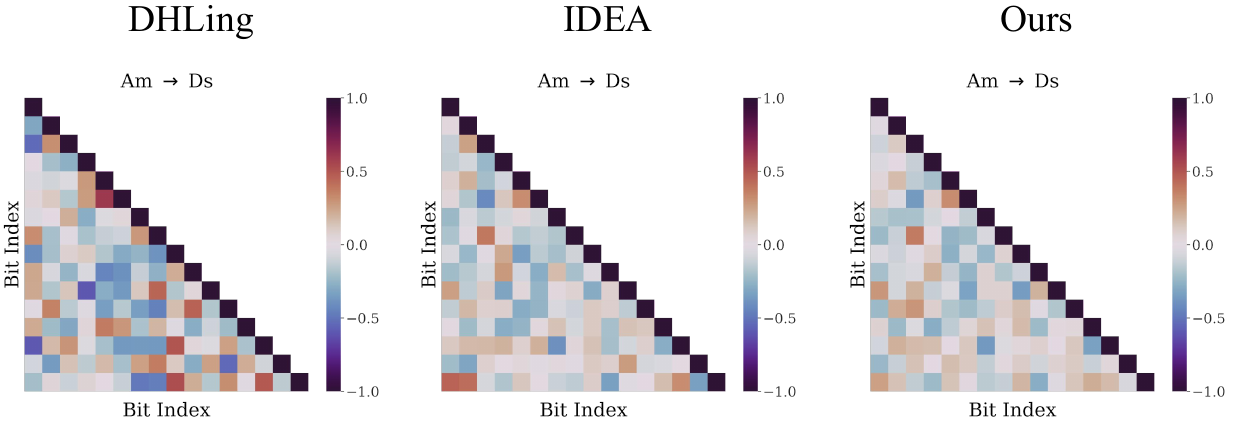}
    \caption{The correlation heatmap of $16$ bit hash codes on Office-31 dataset~(Am $\rightarrow$ Ds).}
    \label{fig:anl-corr-1}
\end{figure}

\begin{figure}[t]
    \centering
    \includegraphics[width=\linewidth]{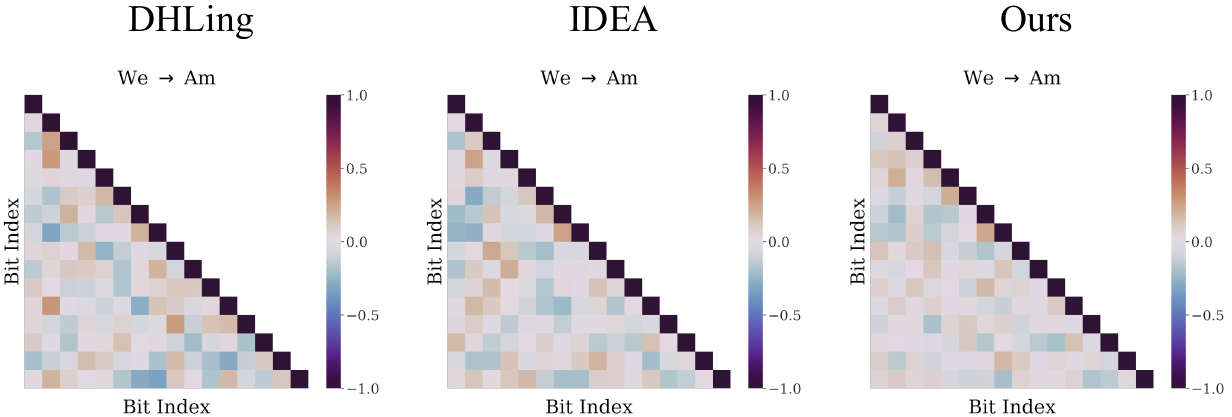}
    \caption{The correlation heatmap of $16$ bit hash codes on Office-31 dataset~(We $\rightarrow$ Am).}
    \label{fig:anl-corr-2}
\end{figure}

\begin{figure}[ht]
    \centering
    \includegraphics[width=\linewidth]{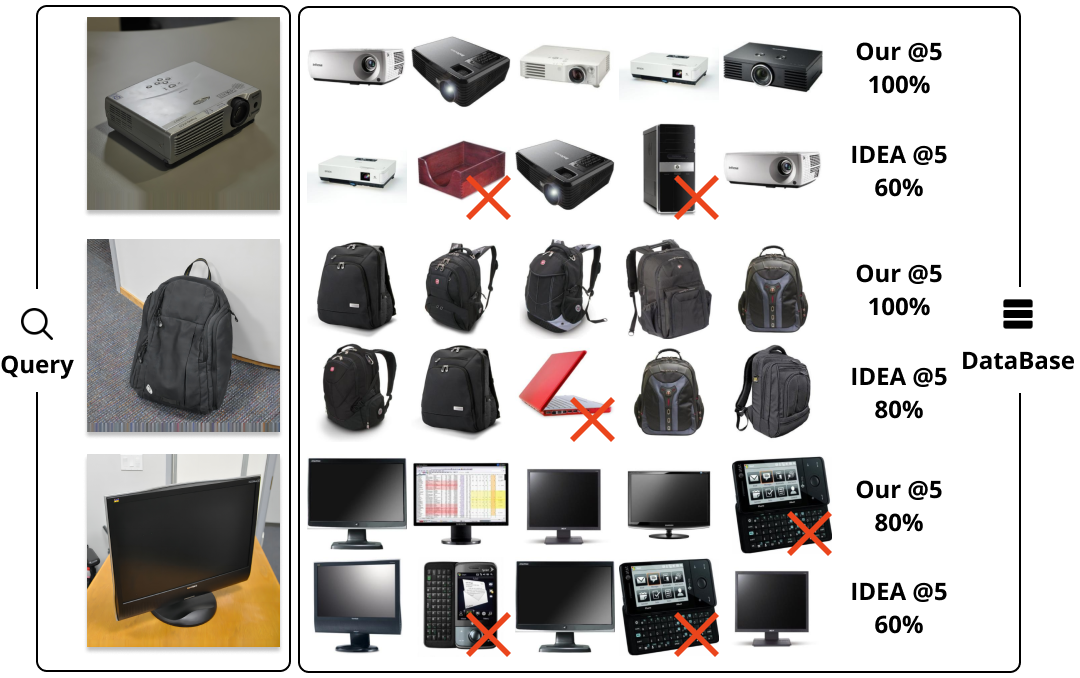}
    \caption{Case study. Query the top $5$ images on the Office-31 dataset.}
    \label{fig:fig6}
\end{figure}

\subsection{Sensitivity Analysis}
\label{sec:sensitivity}

We investigate the proportion $\gamma$ in Eq.~\ref{eq:gamma} and random walk length $k$ in Eq.~\ref{eq:random-walk}, as shown in Figure~\ref{fig:abl-sensitive}. The sensitivity analysis is conducted on Office-Home dataset. 
We first examined the impact of the \textit{early adapter} sample proportion~($\gamma$). Figure~\ref{fig:abl-sensitive} demonstrates that as $\gamma$ increases from $30\%$ to $50\%$, overall accuracy improves, indicating that incorporating more target domain knowledge enhances domain-adaptive hash code learning. However, increasing $\gamma$ from $50\%$ to $70\%$ leads to decreased accuracy, suggesting that excessive target nodes may introduce noise. An optimal $\gamma$ value allows the model to acquire maximal target domain knowledge without introducing excessive noise. Secondly, we explored the impact of different random walk path lengths $k$ ranging from $2$ to $10$. The results are illustrated in Figure~\ref{fig:abl-sensitive}. As $k$ increases from $2$ to $5$, the model performance gradually improves. However, when $k$ further increases to $10$, a slight performance drop is observed. This suggests that increasing the path length can effectively establish progressive cross-domain connections, which also verify our motivation. Nevertheless, excessively long paths may lead to sampling more outliers, thereby affecting the model's generalization ability. Consequently, we set the default $\gamma$ to $50\%$ and $k$ to $5$.

\subsection{Speed Test}
\fix{We conducted speed tests between \method{} and dense vector retrieval on an Intel Xeon CPU E5-2697 v4~($2.30$GHz) with a database of $10^5$ items. We performed $10^3$ runs and report the average retrieval time~(ms) in Table~\ref{tab:time}. The results show that hash codes achieve significantly faster retrieval speeds than dense vectors, demonstrating their efficiency for large-scale retrieval tasks.}

\subsection{Stability Analysis}
\fix{To evaluate the robustness of our method, we conducted stability analysis across five random seeds on the Office-Home dataset. As shown in Figure~\ref{fig:stability}, \method{} not only achieves superior performance but also maintains consistent results, with standard deviations typically below $2\%$. This stability is particularly evident in challenging transfer tasks like Re$\rightarrow$Cl and Pr$\rightarrow$Re, where \method{} demonstrates both higher accuracy and smaller error bars compared to baselines.}

\subsection{Performance under Different Backbones}
\fix{To investigate the impact of different backbones on retrieval performance, we conducted experiments with three representative architectures: VGG-F, ResNet-34, and ViT-Base, as shown in Table~\ref{tab:backbone_comparison}. \method{} consistently outperforms baseline methods across all architectures, with ViT-Base achieving the best performance.}

\subsection{Visualization}
\label{sec:vis}
\paratitle{T-SNE Visualization.}
We utilize t-SNE visualization to demonstrate the discriminative hash codes learned by \method{}, IDEA, and DHLing. As shown in Figure~\ref{fig:tsne}, \method{} can effectively exploit the information capacity of hash codes to learn more discriminative hash codes, thereby achieving more effective image retrieval. \fix{The superior clustering effect of our method can be attributed to two key design aspects: (1) the cross-domain diffusion mechanism effectively filters out noisy samples, leading to more reliable semantic structures, and (2) the hierarchical Mixup strategy progressively aligns domain distributions, resulting in more coherent feature representations across domains.}

\paratitle{Hash Code Correlation Analysis.}
To evaluate \method{}'s ability to learn hash codes with minimal redundancy and maximize hash code capacity utilization, we conducted an analysis using \textit{16}-bit hash codes on the Office-31 dataset. Figure~\ref{fig:anl-corr-1} and ~\ref{fig:anl-corr-2} illustrate the results, with the heatmap representing correlations between different hash bits. Our observations indicate that the proposed method generates more independent hash bits, evidenced by lower correlations between different hash positions. It suggests \method{} can effectively reduce redundancy in the learned hash codes for effective retrieval.

\paratitle{Case Study.} 
We conduct hash retrieval and illustrate the top $5$ retrieval results in Figure~\ref{fig:fig6}. \method{} can achieve higher accuracy in retrieval tasks, thus validating the effectiveness of the proposed method and facilitating downstream tasks based on retrieval.

\section{Conclusion}
\label{sec::conclusion}

This paper investigates a practical yet under-explored problem of adaptive hashing retrieval. We analyze the challenges for the suboptimal performance of existing methods, \ie noise in the target domain and ineffectiveness of direct high-level domain alignment.
We introduce a novel method \method{} from the perspective of graph clustering. \method{}  simulate the domain transfer dynamics by graph flow diffusion and can effectively combat noise to extract reliable samples with lower noise level in the target domain. Leveraging discriminative hash learning, \method{} can stable adapt to the target domain.  
Moreover, \method{} achieves progressive domain alignment with hierarchical Mixup techniques along the cross-domain random walk paths. Through comprehensive experiments in domain adaptive hashing retrieval benchmark dataset with the competing methods, we show the effectiveness of our \method{}. We also conduct visualization and speed test to demonstrate the performance of \method{}.

\paratitle{Limitations. }
\fix{The current \method{} framework is primarily designed for closed-set scenarios, which may limit its applicability in open-set retrieval tasks where target domain contains novel categories.}

\paratitle{Future Work. }
\fix{Future work could explore extending the method to streaming data scenarios, strengthening the theoretical foundations of graph diffusion in domain adaptation, and investigating the integration of foundation models for more robust cross-domain hash code generation.}

\section*{Acknowledgement}

This paper is partially supported by the National Key Research and Development Program of China with Grant No.
2023YFC3341203 as well as the National Natural Science
Foundation of China with Grant Numbers 62276002 and 62306014.

\bibliographystyle{IEEEtran}
\bibliography{7_rec}

\end{document}